# Double-Flow GAN model for the reconstruction of perceived faces from brain activities

Zihao Wang[1], Jing Zhao[2,3], Xuetong Ding[2,3], and Hui Zhang[2,4,5]

[1] School of Computer Science and Engineering, Beihang University, Beijing, China
zihaowang@buaa.edu.cn
[2] School of Engineering Medicine, Beihang University, Beijing, China
[3] School of Biological Science and Medical Engineering, Beihang University, Beijing, China
[4] Key Laboratory of Biomechanics and Mechanobiology, Ministry of Education, Beihang University, Beijing, China
[5] Key Laboratory of Big Data-Based Precision Medicine, Ministry of Industry and Information Technology of the People's Republic of China, Beihang University, Beijing, China
hui.zhang@buaa.edu.cn

**Abstract.** Face plays an important role in human's visual perception, and reconstructing perceived faces from brain activities is challenging because of its difficulty in extracting high-level features and maintaining consistency of multiple face attributes, such as expression, identity, gender, etc. In this study, we proposed a novel reconstruction framework, which we called Double-Flow GAN, that can enhance the capability of discriminator and handle imbalances in images from certain domains that are too easy for generators. We also designed a pretraining process that uses features extracted from images as conditions for making it possible to pretrain the conditional reconstruction model from fMRI in a larger pure image dataset. Moreover, we developed a simple pretrained model for fMRI alignment to alleviate the problem of cross-subject reconstruction due to the variations of brain structure among different subjects. We conducted experiments by using our proposed method and traditional reconstruction models. Results showed that the proposed method is significant at accurately reconstructing multiple face attributes, outperforms the previous reconstruction models, and exhibited state-of-the-art reconstruction abilities.

**Keywords:** Perceived Face Reconstruction, Double-flow GAN, Brain decoding.

## 1 Introduction

Reconstruction of perceived images from brain signals is a hot topic in brain decoding and a prospective part in brain-computer interface. Using functional Magnetic Resonance Imaging (fMRI), a non-invasive neuroimaging technique, researchers are able to measure the neural activities at each brain location across the whole brain when performing particular visual tasks, and therefore makes the perceived objects reconstruction from brain possible. So far, researchers have reconstructed various kinds of perceived objects from fMRI brain activities, such as patterns [34], letters [6, 7, 11], scenes [10,12,15], natural objects [4,16,17].

Of these reconstructed natural objects, face is most special. Face plays a crucial role in our daily life, and the reconstruction of perceived faces is challenging. As compared to common natural object, faces contain more complex high-level features which are hard to define and extract [25]; Faces present multiple face attributes such as identity, race, gender, expression etc., which are holistic and hard to reconstruct [4]. Considering these complex characteristics of face images and the very unique neurocognitive mechanism of faces [xxx], researches have proposed methods specifically for face reconstruction [xx]. In our previous study, we have established a framework using GAN model to reconstruct faces from fMRI signals. Though the method greatly advances the quality of reconstruction for faces, its reconstruction performance is not still satisfactory and needs further improvement.

In this paper, we combined GAN [35] and transformer [33], and proposed a new algorithm framework for high-quality perceived face reconstruction. The framework, which we termed Double-Flow GAN (DFGAN), have three improvements compared to the previous GAN-based reconstruction models. First, Given the shortage of perceived face fMRI dataset, our framework was divided into two stages. In the first stage, the DFGAN was pretrained in a large dataset which contains only face images. In the second stage, the features extracted from face-selective brain regions were aligned with face image features, and used as conditional inputs to the GAN model. Second, the pretraining framework alleviated the difficulty of fitting complex models arising from the paucity of fMRI data, so models with large parameters and high level of expressive ability such as Transformer can be used as basic models in GANs to reach a higher resolution reconstruction. Third, the discriminator of the GAN was enhanced by taking input from both positive and negative samples in a single forward process, which allows it to compare the two samples and ultimately decide which one is true. It could supervise generator more so that the generator continues to optimize when similar but not exactly correct faces have been generated.

## 2 Related Works

Many methods were used for perceived face reconstruction. Principle component analysis (PCA) was widely used in early years, for its ability to provide a compact representation of facial features by extracting a set of orthogonal basis vectors, known as eigenfaces [34]. Researchers could simplify the task of reconstructing face images with detailed attributes to the task of reconstructing low-dimensional feature vectors. As a result, combining PCA and basic supervised machine learning methods like support vector machine (SVM) and linear regression, or further extracted features to supervise can help to achieve reconstruction [4, 27, 29, 30]. However, reconstruction based on feature extracted by PCA lose detailed information and the reconstructed images are very blurry.

In recent years, deep generative networks have shown great capability in image synthesis. Convolution neural networks (CNN) was a widely used method for feature extraction and generation. Conditional generative adversarial networks (cGAN) provides a direct way for reconstructing face images from brain signals. The idea of generative adversarial networks (GAN) is to train two neural networks, where the two networks engage in a competitive game that generator tries to generate images to fool the discriminator and discriminator tries to correctly classify real and fake images. Works combining dimensionality reduction method with CNN-based deep generative network have shown great abilities generating faces with high-resolution [32]. In addition, modifying the loss function of the GAN by adding attribute losses is also a good way to improve the consistency of reconstructed face attributes [4]. However, the discriminator in traditional GANs can only predict the truth or falsity of an image, whereas all faces are similar and differ only in details, which makes the generator's task easy and the discriminator's task difficult, which will harm the game between them and the reconstruction work. Moreover, the representation capability of CNN-based generators is not sufficient to generate high-resolution images. Moreover, masked auto-encoder (MAE), by predicting masked pixels through decoder, performed well in feature representation and reconstruction [36]. Finally, diffusion models (DM), which have been gaining attention and used in high-resolution natural-image reconstruction [10,24], were not applicable for perceived face reconstruction currently, because DMs focus on semantic contents reconstruction and are less of consistency, and minor changes in pixels of face images will cause changes in attributes.

Our proposed perceived face reconstruction framework DFGAN focused on the question of insufficient representation capability of cGAN and the imbalance between generator and discriminator. Our main contributions are as follows: 1) We proposed a novel GAN discriminator structure that can handle imbalances in images from certain domains that are too easy for generators. 2) We designed a pre-training process using features extracted from images as conditions that can be used in domains where neural data is lacking.

## 3 Method

### 3.1 Overview of the reconstruction framework

We propose an algorithmic framework that can accurately reconstruct multi-attribute face images from fMRI brain signals containing multiple brain regions as shown in Figure 1. Our framework consists of three modules: a multi-task feature extraction network, a liner model and a dual transformer-based generative adversarial network. The multi-task feature extraction network is used to extracting multi-attribute features from facial images. The liner model aligns fMRI data to the feature of face image to alleviate the influence of differences between different individual’s brain. The dual transformer-based generative adversarial network is the face generation module, which is used to reconstruct the perceived face images from the multi-dimensional face features predicted by the brain signals.

The whole training process are divided into two steps, 1) In pre-training process, we use the VGG-face model to extract features from the dataset, and then pre-train the generative model to construct the relationship between the features and the face image. 2) In fine-tuning of the model, we constructed a multi-task face feature extraction model based on the VGG-face model in the pre-training to extract three facial features, expression, identity and gender feature. The linear model performs a linear mapping of fMRI-extracted signals from multiple brain regions to three facial attribute features. The fmri signals are mapped to the three face features and used as inputs to the generative network, and the face images are trained as outputs of the generative network. After the entire model framework is pre-trained, we use the collected data to fine-tune the generative network.

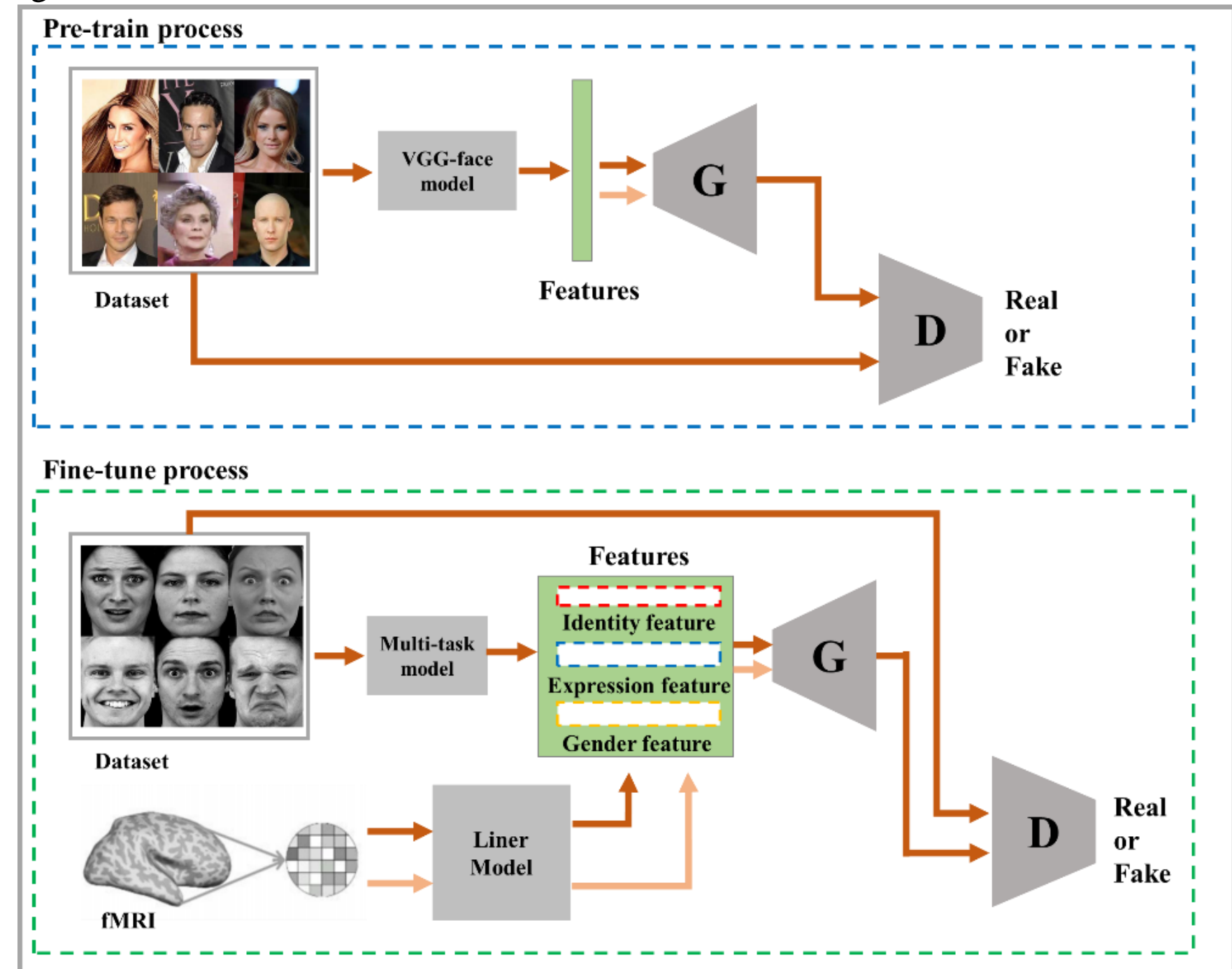

Fig.1. Overview of the whole approach

### 3.2 A multi-task feature extraction network

To extract the face features representing multiple attributes of facial expression, identity and gender without bias, we used a multi-task network [4] in our study. The network has one input of a single face image, and three outputs to identify the face image's expression, identity, and gender category, respectively. The multi-task network based on VGG-face shares parameters in first 14 layers and then divides into three output branches. The three branches have the same network layers, with five convolutional layers, five maximal pooling layers, and three fully connected layers in that order. The network parameters for these three branches were determined by fine-tuning the training. We replaced its first two fully-connected layers from 4,096 to 512 dimensions, and defined the last fully-connected layer of each output as 7, N and 2 dimensions, respectively. Here, 7 represents the seven basic facial expressions: fear, anger, disgust, happiness, neutral, sadness, and surprise, N is the number of facial identities for re-training the model, 2 represents the gender categories of male and female.

### 3.3 The liner model

The brain structure differs among people, there is a big difference between fMRI signals collected from different individuals. Therefore, it's hard to synthesize faces directly through fMRI data collected from different people. We used a set of linear regression models to establish the mappings between the brain signals and the multi-dimensional face features. In our study, we padded the data to the same size and tried to learn a linear model that could project fMRI to the feature of face images. The linear model could project fMRI from different people to the same feature to alleviate individual variation in the reconstruction process.

### 3.4 Double-flow Transformer GAN

To reconstruct high-quality face images, we proposed a Double-flow Transformer GAN (DFGAN) to realize more precise face image reconstruction with the desired facial attributes. DFGAN consists of a transformer-based generator and a dual-flow transformer-based discriminator.

**Swin Transformer-based Generator** To reduce the large computational complexity of the global attention, we utilized Swin Transformer instead of Transformer block, which processes the entire image into several separate non-overlapping patches called "windows." The Swin Transformer consists of six stages, each composed of two Transformer blocks. Each block contains two sub-layers: a shifted window-based self-attention mechanism and a feed-forward neural network. The shifted window-based self-attention mechanism enables the model to capture long-range dependencies within the image by shifting the windows and attending to neighboring windows. Specifically, the generator takes the one-dimension face feature $z$ as its input and passes it through an embedding $x = \tau_\theta(z) \in R^{B^2 \times D_{emb}}$, where $B$ representing the bottom-width and $D_{emb}$ represents the dimension of embedding. Then, $x$ could be seen as an $B \times B$ image with $D_{emb}$ channels and we can use $x$ as the input of Swin Transformer . Considering both high-resolution image generation and computation cost, we use bicubic up sample at early stage and pixel shuffle at latter stage, with stage blocks using Swin Transformer between them. The detailed generative model is showed in **Fig. 1**.

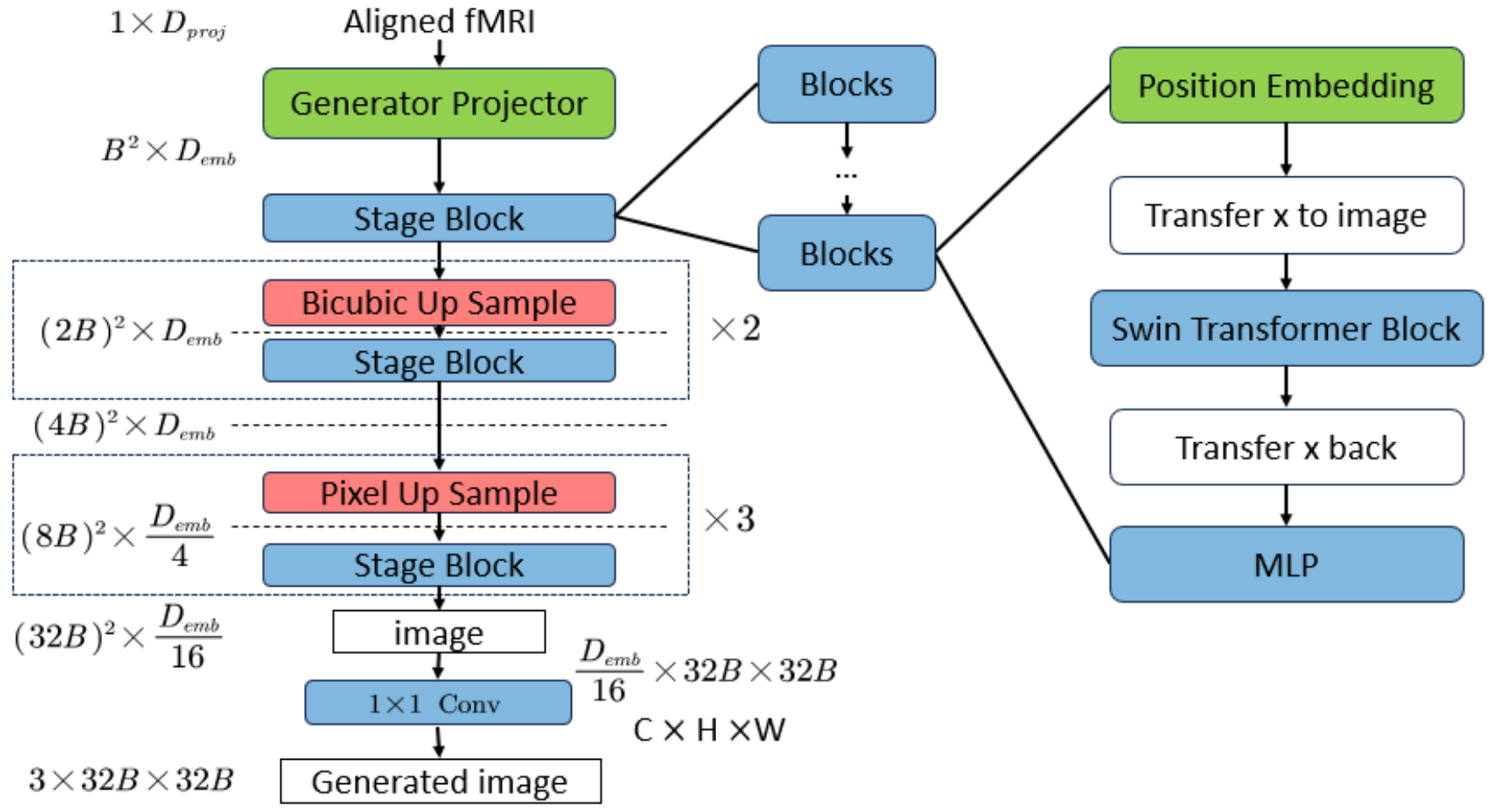

**Fig. 1.** The architecture of our generator based on Swin Transformer.

**Double-flow Transformer Discriminator** To increase the ability of discriminator and alleviate the imbalance between the tasks of generator and discriminator in this situation, we introduced Double-flow Transformer Discriminator, real-fake image pairs are input into the model, and the output of model is a pair of numbers between 0 to 1 that are added up to one, representing the probability that image 1 or 2 is true.

As shown in **Fig.2**, the discriminator contains two major parts: the feature extraction and prediction. In the feature extraction part, images are divided into separate patches, then these patches are embedded and inputted to a transformer encoder block. The transformer encoder contains four blocks, which consist of a self-attention layer, linear model and a feed-forward network. In the prediction part, features will be processed by a comparison module to focus on the differences between real and generated faces. Also, we modified the output structure of the last linear layer of discriminator and added a task, which not only proves whether the image is real or fake, but also predicts the identity, gender and expression of the face, respectively.

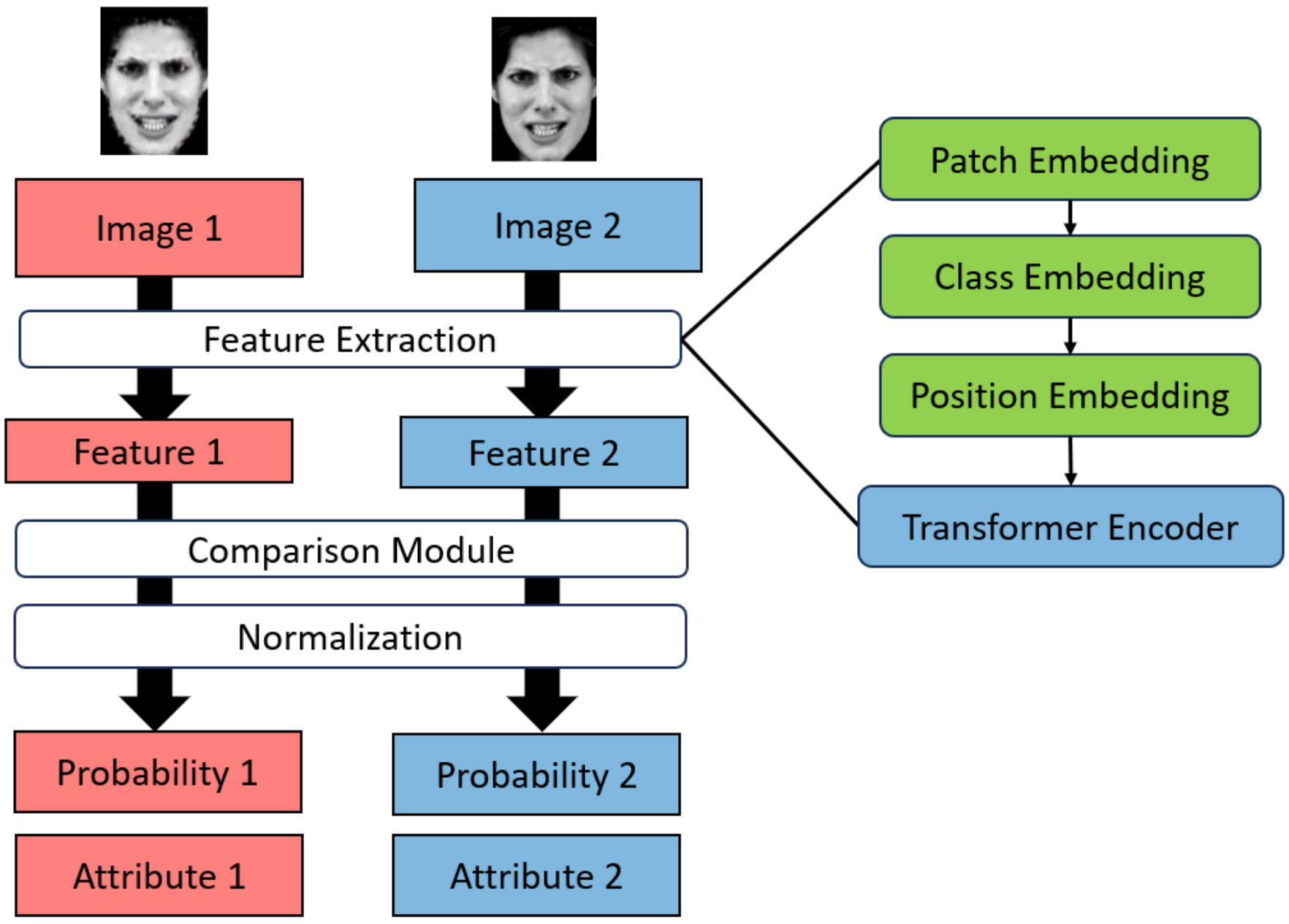

**Fig 2.** Architecture of discriminator

**Comparison Module** Faces contain multiple similar features in terms of contour, facial distribution, etc. Therefore, without constraints such as gender, expression, and identity, the images generated by the generator can mislead the classification results of the discriminator, making it difficult to achieve a balance in adversarial training between them.-And the optimization of generator will stuck here.
Also, the consistency of the reconstructed images is important when generating them from fMRI that a person can only perceive single face at one time. In order to drive the generator to reconstruct images with high quality and consistency, we introduced two methods to make the discriminator powerful. More information and reasonable prior knowledge provided always make the model easier to learn and converge faster. We decided to design a discriminator which takes both generated fake image and the true image as the input without knowing which is the true, and the task is to distinguish which image is true. So the discriminator can "compare" two images in one process and use information from both images to decide which is better. In order to realize the compare function, we designed the so-called Comparison Module.

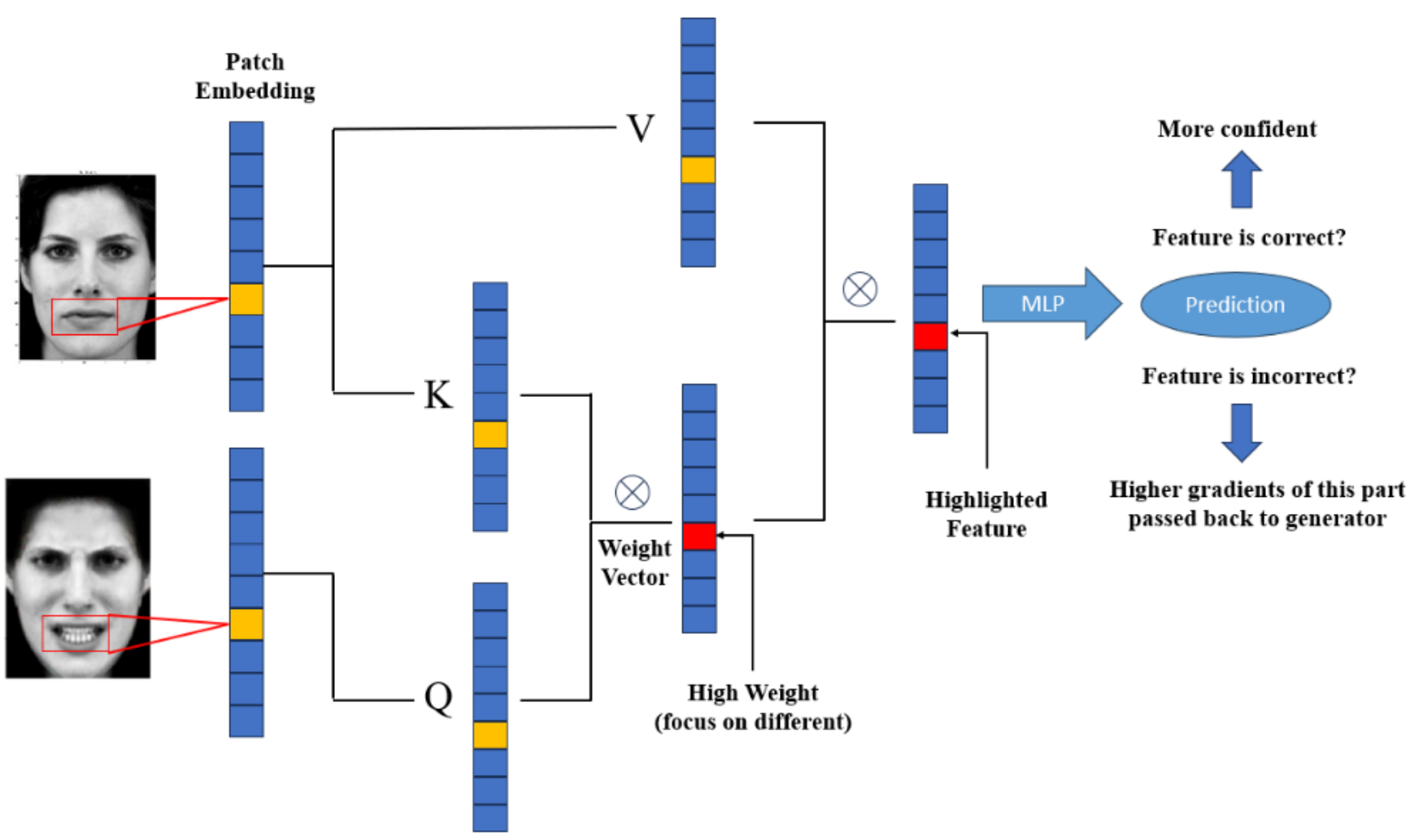

**Fig. 2.** The structure of comparison module. It highlights different parts between two input images, and help generator to fine-tune the generation of feature

The main part of Comparison Module is Cross-Attention [3]. Let $l_1 \in R^{B\times k}$and $l_2 \in R^{B\times q}$ be the representations of Image 1 and Image 2 respectively. Using $l_1$ as the value and $l_2$ as the key value, the model will calculate the probability that $l_1$ is the feature representation of the true image given that the feature representation $l_2$ of another image is known. The equation of single head Cross-Attention is:

$$Q = l_2 \times Q^{q\times q}$$
$$K = l_1 \times K^{k\times q}$$
$$V = l_1 \times V^{k\times q}$$
$$Output = Attention(Q, K, V) = softmax\left(\frac{QK}{\sqrt{q}}\right)V$$

in our implementation, number q and k are the same. The equation for the other image is similar.

To effectively utilize the attention module, we used class embedding and position embedding before this part, and a linear layer after the part to map the output to final prediction. Details are shown in **Fig. 2**.

Unlike the usage of Cross-Attention in multimodal tasks, which are used to incorporate information from one modality to another, our paper only uses it as a scaler to scale the image vector. This method uses the information from another image to calculate the attention score of the processing image, enabling the model to "focus" on major difference between two images and avoid the influence of common places like backgrounds. Going back to faces, when the generator can produce medium-quality faces, it is likely to confuse a normal discriminator, but a discriminator with this module can observe a more perfect image of a real face and focus on the imperfect details, then negate the fake image and force the generator to optimize further. In addition, the module can also act as a pixel loss to allow the model to converge faster, for example, when the image generated at the beginning is noisy, the discriminator can obtain the real image and pass back more information.

### 3.5 loss function

As a result, we could further modify the loss function, and the loss function of the discriminator is defined as:

$$\begin{aligned} &\max_D L_{GAN}^{D} = E_{x\sim Px(x)}[logD(x)] + E_{z\sim Pz(z)}[log(1-D(G(z)))] \\ &-\alpha \times (L_{BCE}(t_{attr}, D_{attr}(x)) + D(G(z)) \times L_{BCE}(t_{attr}, D_{attr}(G(z)))) \end{aligned} \tag{1}$$

where x is the real face images, z is the aligned latent of fMRI, $t_{attr}$ is a concatenation of features of face identity, expression and gender, $L_{BCE}$ is the binary cross-entropy loss and $D(x)$ is the possibility that discriminator predicted the image real or fake. Compared with the loss function proposed in [4], we added a factor of confidence level to the attribute loss, with the insight that the discriminator has less responsible for the predicted attributes of the image which is predicted fake.

According to the discriminator loss, the generator loss has the same structure, which is defined as:

$$\begin{aligned} \min_G L_{GAN}^{G} = &[log(1-D(G(z)))] + \lambda L_{MAE}(G(z), x) \\ &+ \alpha D(G(z)) \times L_{BCE}(t_{attr}, D_{attr}(G(z))) \end{aligned} \tag{2}$$

where $L_{MAE}$ is the MAE loss and the other annotations are the same as the equation shown above. Compared with the loss function proposed in [4], we added an attribute loss for the generator with insight that the generator should have responsible for attributes of the generated images that are sufficient to confuse the discriminator.

## 4 Experiments

### 4.1 Dataset

We used the CelebA dataset for the pre-train part of the model [5]. The dataset contains 30,000 of face images. We only use the face images in the dataset, and each image was resized to 224 × 224 to meet the requirement of input of VGG Face. Also, the images were resized to 128 × 128 before put into the discriminator in order to meet the shape of reconstructed images.

Moreover, we used the fMRI-face pair dataset introduced in the previous paper [4]. Briefly, the face part of dataset consists 952 frontal face images, which contain 136 different identities, 7 different facial expressions, and 60 of the identities are female. These faces come from KDEF dataset [37] and RaFD dataset [38]. and then they are converted to grayscale images and normalized to have the same size, brightness, and contrast to minimize low-level visual differences. Also, the dataset contains 2,800 pairs of fMRI-image acquired from the OFA, amygdala, STS, FFA, and aIT brain regions of 2 individuals, each containing 1,400 pairs divided into 1,260 pairs for training and 140 pairs for evaluation. We used the fMRI data which has been processed by the MTDLN introduced in the same paper, which mapped the original to a feature space under supervision of the identity, gender and expression attribute. We used the processed fMRI as the input of fMRI projector introduced in 3.1 and utilized the attributes as attribute loss in 3.3. More details about the dataset are provided in the original paper.

### 4.2 implementation

We implemented our model in PyTorch with one NVIDIA GeForce RTX 4090 GPU which has 25GB of memory. Our model is trained using the Adam optimizer with a learning rate of 2e-4, for 200 epochs in pre-train process and 500 epochs in fine-tune process. As for the generator, dimension of condition is 128, depth of transformer blocks is 2, number of transformer blocks is 5, number of heads of each attention is 4, window size is 16, and dropout rate is 0. As for the discriminator, patch size is 8, token dimension is 128, depth of transformer blocks is 4, number of heads of each attention is 4 and dropout rate is 0.1. Moreover, the parameters $\lambda$ and $\alpha$ is empirically set to 10 and 0.01. All other hyper-parameters are retrained using their default settings. Competing methods

We compared our model to the method used in the original paper [4]. The baseline model is mcGAN, and used a simpler loss function:

$$
\begin{aligned}
\max_{D} L_{GAN}^{D} = & E_{x\sim P_d(x)}[log D(x)] + E_{z\sim P_z(z)}[log(1-D(G(z)))] \\
& - \lambda_D \{ L_{BCE}(t_{id}, D_{class_{id}}(x)) + L_{BCE}(t_{id}, D_{class_{id}}(G(z))) \\
& + L_{BCE}(t_{exp}, D_{class_{exp}}(x)) + L_{BCE}(t_{exp}, D_{class_{exp}}(G(z))) \\
& + L_{BCE}(t_{gen}, D_{class_{gen}}(x)) + L_{BCE}(t_{gen}, D_{class_{gen}}(G(z))) \}
\end{aligned} \tag{3}
$$

$$
\min_{G} L_{GAN}^{G} = E_{z\sim P_z(z)}[log(1-D(G(z)))] + \lambda_G L_{MAE}(G(z), x) \tag{4}
$$

where notations are the same as the origin paper.

### 4.3 Quantitative assessments

We use mean square error (MSE), structure similarity index measure (SSIM) and attribute error to assess our reconstruction model. MSE measures the pixelwise difference between real faces and reconstructed faces. SSIM takes into account the brightness, contrast and structure of the image, and measures the similarity of faces by the average gray value, gray standard deviation and correlation coefficient.

The attribute error is intended to be used for measure the consistency of reconstructed attributes, and is defined as the mean square error between predicted attributes of reconstructed faces and real attributes. In order to obtain attributes of reconstructed faces, we first trained a ResNet-50 by predicting attributes of real faces and used it as the attributes predictor of reconstructed faces. The equation is:

$$L_{attri} = \sum \left( R_{attri} - P_{attri}\left(G(z)\right) \right)$$

Where $R_{attri}$ isreal attributes value, $P_{attri}$ is predictor, G is generator and z is the fMRI data.

# 5 Results

## 5.1 Visualize results

It is first necessary to verify the effectiveness of the comparison module as well as the models proposed in this paper. We will compare the effect of the following classical models under the same conditions: mcGAN, for the conditional GAN using multi-attribute loss, which is the baseline and the generator and discriminator in the GAN are convolutional neural network; mcGAN+comparison module (mcGAN+compare), that is, add the comparison module on mcGAN, which is implemented using the output of the last fully-connected layer of mcGAN discriminator as the feature of the two pictures and the input of the comparison module, and the feature is used to do cross-attention; mcGAN+self-attention (mcGAN+onlyAttn), self-attention is added in the same position with the former comparison module, and the feature of each picture is done with itself to do the self-attention to ensure that the number of the added parameter is the same as the model of adding the comparison module; using convolutional neural network generator with Transformer-based discriminator (CNNG+TD), whose discriminator structure is the same as that of this paper's discriminator but removed the comparison module; using CNN's generator with the discriminator proposed in this paper (CNNG+dfD); the generator used in this paper with the Transformer-based discriminator , whose discriminator structure is the same as the discriminator removal comparison module in this paper (TG+TD); our proposed model. All of these models have exactly the same training parameters, training time and loss function design except for the different structure.

**Fig. 3** shows the reconstruction effect of different models on the first 8 images in the test set. It can be seen that our proposed method can achieve facial image reconstruction and accurately reconstruct three features: expression, identity, and gender. Also, the **Fig. 4** shows the samples of ablation studies, which performs the role of alignment, modification of loss function and pretrain.

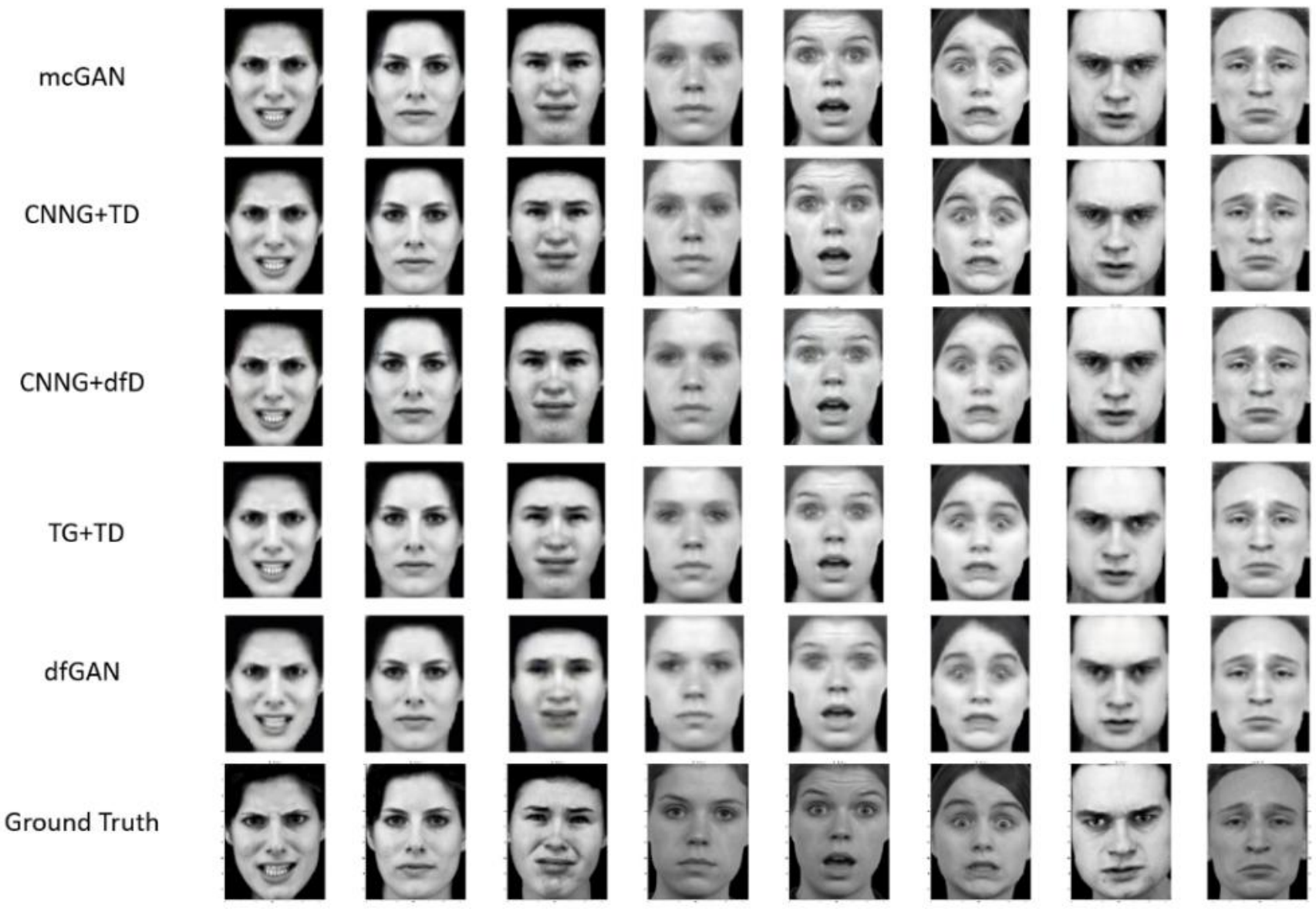

**Fig. 3.** Samples of the generated images

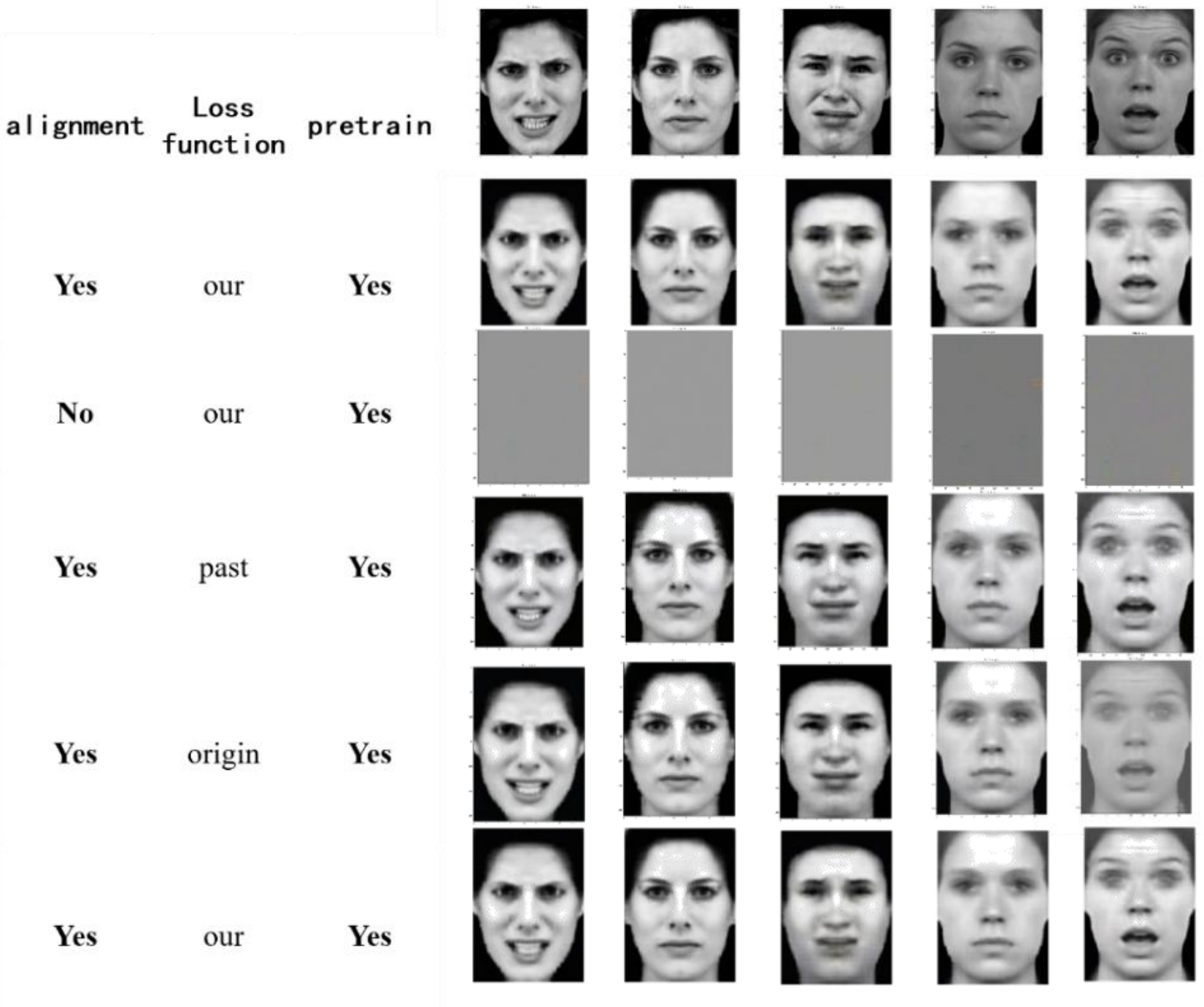

**Fig. 4.** Samples of the ablation experiments

### 5.2 Quantitative results

**Table 1** shows the results of all the experiments, firstly, comparing mcGAN, mcGAN+comparison module and mcGAN+self-attention, it can be seen that increasing the comparison module or self-attention as a method to improve the model expression ability and the number of parameters can improve the model effect, but when increasing the same number of parameters, comparison module relative to the self-attention improves the model effect more significant, respectively, make the MSE of the baseline model decreased by 41.3% and the SSIM increased by 16.3%, while the self-attention only decreased the MSE by 15.2% and increased the SSIM by 11.1%. This result validates the effectiveness of the comparison module.

**Table 1.** Quantitative results on dataset using different models, in terms of SSIM, MSE and Attribute error. The attribute error is computed from a pre-trained ResNet 50 that can predict the attribute scores of the real image, where the loss is between the predicted attribute scores of the generated image and the real attributes. The best results are marked in red. All fMRI data (contains baseline) are projected. T represents Transformer and dfD represents our double-flow Discriminator.

| Model | MSE | SSIM | Attribute error |
|---|---|---|---|
| Baseline (mcGAN) | 0.046 | 0.578 | 0.932 |
| mcGAN+compare | 0.027 | 0.672 | 0.937 |
| mcGAN+onlyAttn | 0.039 | 0.642 | 0.939 |
| CNNG+TD | 0.027 | 0.689 | 0.899 |
| CNNG+dfD | 0.028 | 0.669 | 0.907 |
| TG+TD | 0.027 | 0.695 | 0.900 |
| proposed model (dfGAN) | 0.025 | 0.695 | 0.899 |

The results show that improvements can be obtained using either the Transformer-based generator or the discriminator. Especially, our proposed discriminator is better than Transformer discriminator when using Transformer generator, but weaker than it when using CNN generator. We think this might be because our enhanced discriminator achieved a good balance with the Transformer generator but was too strong for the CNN generator.

**Table 2.** Quantitative results of ablation experiments, in terms of SSIM and MSE. The base model is dfGAN. The best results are marked in red.

| align | loss function | pretrain | MSE | SSIM | Attribute error |
|---|---|---|---|---|---|
| yes | our | yes | 0.025 | 0.695 | 0.899 |
| no | our | yes | 9.235 | 0.259 | 9.891 |
| yes | past | yes | 0.027 | 0.679 | 0.910 |
| yes | origin | yes | 0.031 | 0.667 | 0.904 |
| yes | our | no | 0.309 | 0.608 | 0.926 |

**Table 2** and **Fig. 4** shows the results of the ablation experiments, where the first row of **Fig. 4** are real images. In **Table 2**, "our" refers to the loss function of this paper, "past" refers to the loss function without confidence parameters used, and "origin" refers to the loss function without attribute loss used. The analysis is as follows.

Comparing whether the alignment is used or not, it can be seen that with the loss function and pre-training, not using the alignment leads to the model not being able to be work, with its MSE increasing by a factor of 368.4 compared to the proposed model, the attribute loss enlarging to a factor of 11, the SSIM decreasing to 37.3, and the sample in Fig. 4 being as messy as it should be. The reason this remains consistent with the previous speculation of alignment and pre-training is due to the fact that the generative model is relatively familiar with the data in the vicinity of the image features extracted by the same model after pre-training, whereas re-fitting the fMRI mapping would lead to difficult learning if it were to be re-fitted again. The effect on models that do not use alignment and also do not use pre-training will be discussed further in Section 5.4 Generalization Capability.

Comparing which loss function is used, it can be seen that using the original loss function without confidence is 8% higher than the improved loss function in this paper in terms of MSE, a 2.24% decrease in SSIM, and a 1.22% increase in attribute loss, while using the original loss function is 24% higher in terms of MSE, a 4.03% decrease in SSIM, and a 0.56% increase in attribute loss. From this, it can be concluded that the loss function in this paper is optimal in all three major indicators.

Comparing whether pre-training is used or not, the model that also uses the alignment and the loss function of this paper but does not use pre-training has an increase of 12% in MSE, an increase of 8.3% in attribute loss, and a decrease of 4.4% in SSIM compared to the model of this paper, and it can be found that the semantic information of its reconstructed face is preserved, and there is no obvious problem in the presentation of the generated image, and the increase of attribute loss is within the acceptable range.

### 5.3 Sensitivity Analysis

There are two hyperparameters that are very important in the reconstruction framework, the factor of attribute loss $\alpha$ and the factor of pixel loss $\lambda$. these two parameters affect the significant proportion of the reconstructed face positive error, the reconstructed attribute difference, and the pixel-level difference of the reconstructed image. In this experiment, firstly, $\lambda = 10$ was fixed, and $\alpha$ was adjusted by orders of magnitude, 0.001, 0.005, 0.01, 0.02, 0.05, 0.1, 0.5, 1, 10, from which $\alpha$ of the best model was selected, and after that $\alpha$ was adjusted by fixing $\alpha$ to 1, 5, 10, 20, 50, 100. At the same time, this experiment will also calculate the misjudgement rate (misjudge) of the discriminative model for the generated images on the validation set, i.e., how many of the generated images are misjudged as real images, as a way to validate the motivation of this paper - the importance of the equalization of the generative discriminative power on the effectiveness of the model.

**Table 3.** Sensitivity analysis of fixed $\boldsymbol{\lambda = 10}$

| α | MSE | SSIM | misjudge |
|---|---|---|---|
| 0.001 | 0.03 | 0.682 | 3.90E-07 |
| 0.005 | 0.028 | 0.683 | 0.006657 |
| 0.01 | 0.027 | 0.691 | 0.578 |
| 0.02 | 0.026 | 0.689 | 0.999 |
| 0.05 | 0.03 | 0.676 | 0.99 |
| 0.1 | 0.027 | 0.690 | 0.99 |
| 0.5 | 0.028 | 0.680 | 0.99 |
| 1 | 0.03 | 0.683 | 0.99 |
| 10 | 0.026 | 0.686 | 0.99 |

**Table 4.** Sensitivity analysis of fixed $\boldsymbol{\alpha = 0.01}$

| λ | MSE | SSIM | misjudge |
|---|---|---|---|
| 1 | 0.025 | 0.695 | 0.599 |
| 5 | 0.028 | 0.646 | 0.602 |
| 10 | 0.027 | 0.691 | 0.578 |
| 20 | 0.029 | 0.676 | 0.599 |
| 50 | 0.028 | 0.685 | 0.578 |
| 100 | 0.026 | 0.695 | 0.506 |

Firstly, a combination of **Fig. 5**, **Table 3**, and **Table 4** shows that sharp changes in α and λ do not significantly affect the model effect, which is maintained even when there are orders of magnitude changes in the two hyper-parameters (α spanning 4 orders of magnitude and λ spanning 2 orders of magnitude). The worst results in the entire experiment were MSE = 0.03 and SSIM = 0.646, which is only a 20% increase in MSE and a 7.1% decrease in SSIM relative to the best results. This shows that the model is not very sensitive to hyperparameters.

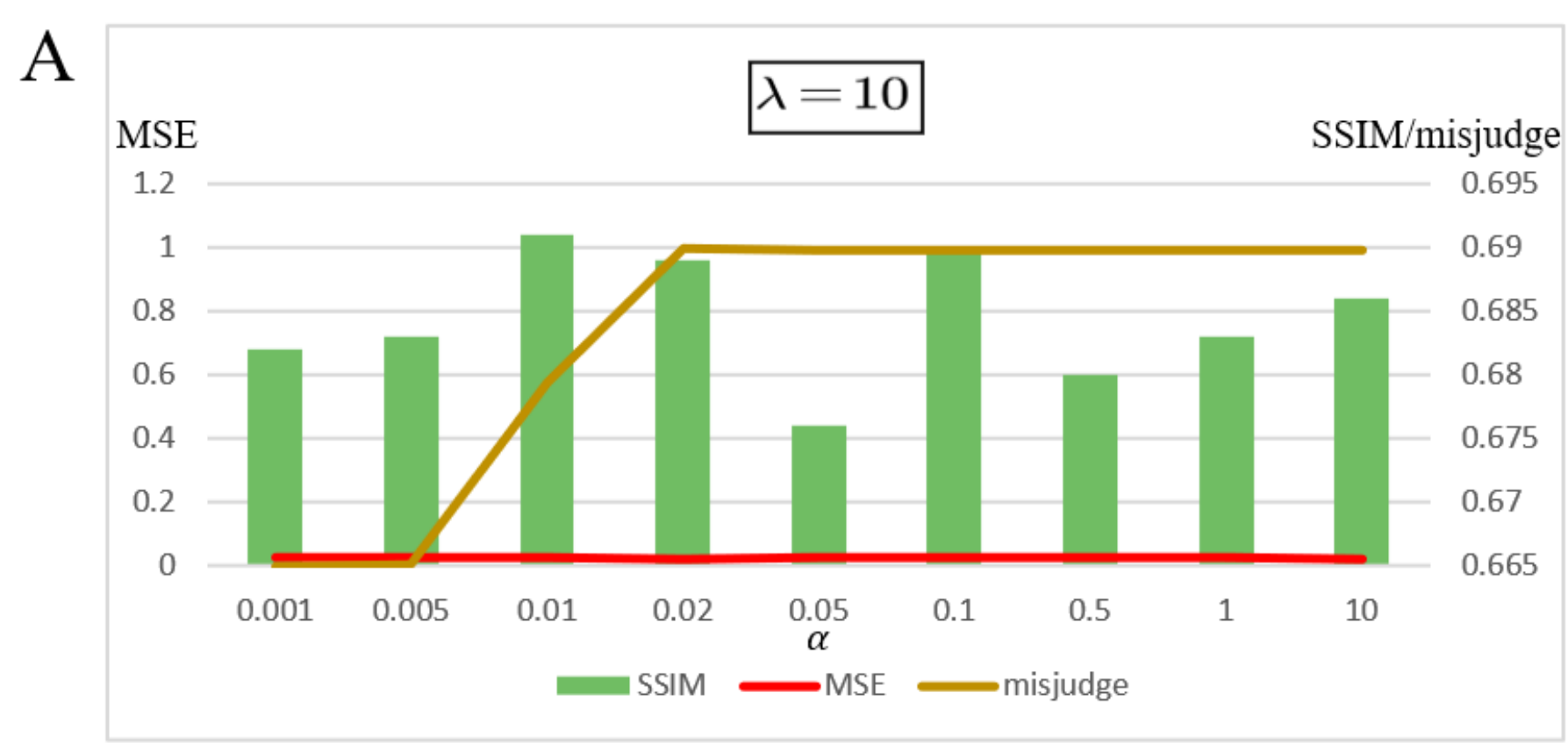

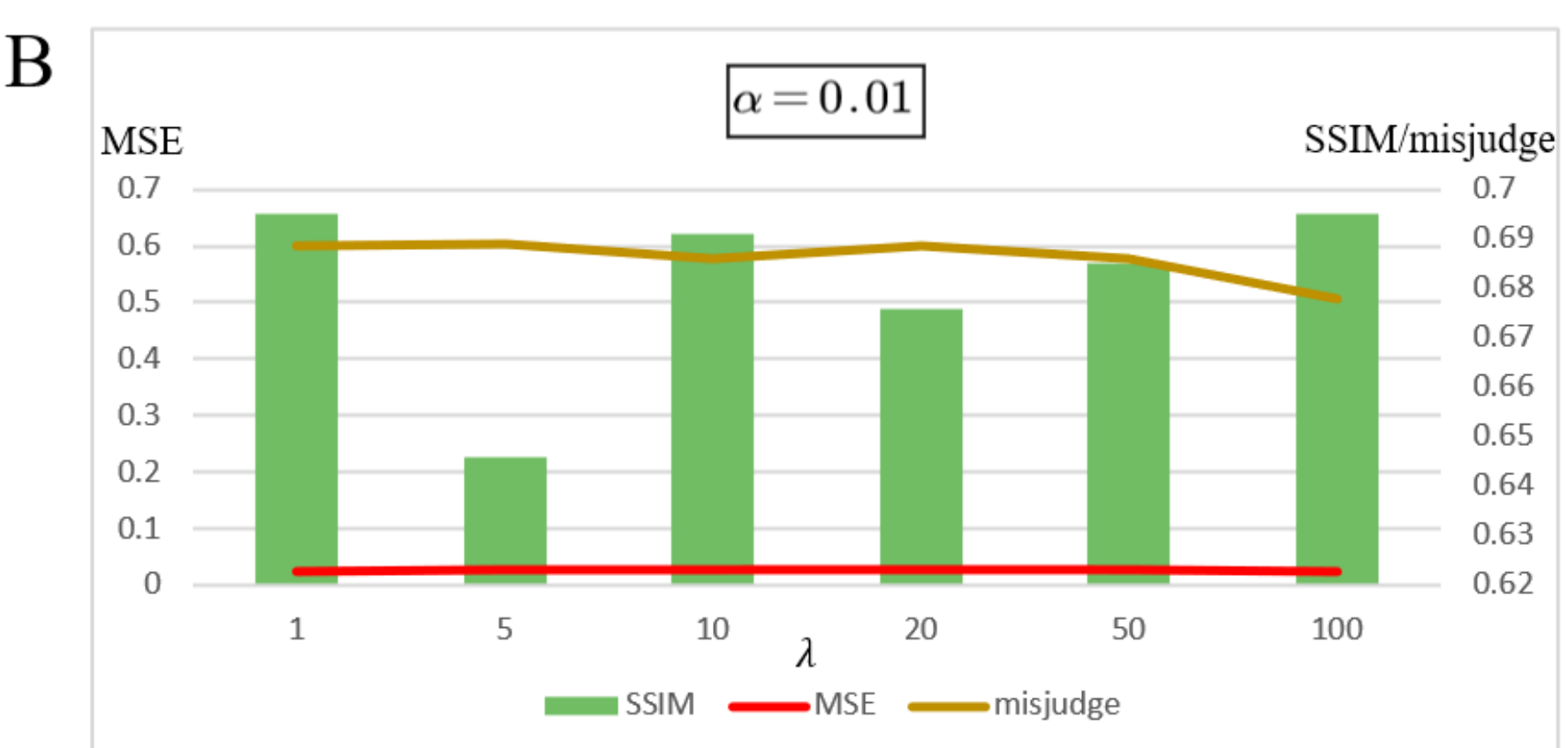

**Fig. 5.** Results of sensitivity analysis. $\alpha$ is the horizontal coordinate of A, and $\lambda$ is the horizontal coordinate of B. For both images, misjudge and MSE are left vertical coordinate, SSIM is right vertical coordinate. The score of misjudge represents what percentage of false images the discriminator will determine as true.

For the generative discriminative ability difference, based on **Fig. 5 A**, SSIM reaches its highest value when α is 0.01 and the wrong score rate is 0.578, which is in the middle of 0 and 1, which matches the motivation of this paper to adjust the discriminator that the GAN will achieve the best results when the generator is balanced with the discriminator. Also as shown in **Fig. 5 B**, when fixing $\alpha = 0.1$ to change the λ, the misclassification rate fluctuates between about 0.5 and 0.6, which demonstrates that the generator and the discriminator continue to compete in this interval.

### 5.4 Generalization Capability

Due to the significant variation in fMRI data extracted from different individuals, there is a challenge in achieving generalization capability with reconstruction models. In order to address this issue, we tested to determine if the reconstruction model could accurately reconstruct perceived images for new individuals. Our dataset contains fMRI data from two person. We used data from one person for training, and evaluated the model's performance on the fMRI data from the other person for testing.

**Table 5.** Quantitaive results of the Inter-subject experiments

| test | all | | 1 | | 2 | |
|---|---|---|---|---|---|---|
| train | MSE | SSIM | MSE | SSIM | MSE | SSIM |
| all | 0.025 | 0.695 | 0.026 | 0.695 | 0.024 | 0.695 |
| 1 | 0.16 | 0.638 | 0.159 | 0.645 | 0.161 | 0.632 |
| 2 | 0.03 | 0.672 | 0.031 | 0.669 | 0.029 | 0.677 |

As shown in **Table 5**, the results of training under all fMRI dataset (all), Individual 1, and Individual 2's fMRI data, and predicting under all fMRI dataset (all), Individual 1, and Individual 2, respectively, are tested. It can be seen that the worst data MSE = 0.16 and SSIM = 0.632, whose decrease in effect is less pronounced compared to training on all data, suggesting that the model generalizes well across individuals. Meanwhile, we tested the effectiveness and generalizability of models that do not use both alignment and pre-training as follows.

**Table 6.** Quantitive results of the Inter-subject experiments with no pretrain and alignments

| test | all | | 1 | | 2 | |
|---|---|---|---|---|---|---|
| train | MSE | SSIM | MSE | MSE | SSIM | MSE |
| all | 0.028 | 0.692 | 0.028 | 0.028 | 0.692 | 0.028 |
| 1 | 0.09 | 0.48 | 0.026 | 0.09 | 0.48 | 0.026 |
| 2 | 0.136 | 0.46 | 0.195 | 0.136 | 0.46 | 0.195 |

As shown in Table 6, the model without alignment and pre-training was able to successfully learn the face reconstruction method on the full training set, but the results are still lower than the model in this paper. Meanwhile, its generalization is significantly lower than this paper's model, with the worst MSE reaching 0.195 and SSIM reaching 0.277.

In order to demonstrate the generalizability brought by alignment and pre-training more intuitively, the first and the last images in the test set are selected for partial visualization in this paper.

Alignment and Pretrain | train | test 1 | test 2 | Ground truth

Yes 1

2

No 1

2

Fig. 6. Visualization of Generation

As shown in **Fig. 6**, this experiment visualizes the first and last images of the test set, while the contrast is increased by 50% for the faces in the first and fourth rows in order to observe them more clearly. It can be seen that when using alignment and pre-training, although the generated images are lighter in tone and relatively blurred when using Individual 1 as the training set, it can be found that the generated images can still tell the general appearance of the face and distinguish the gender and expression, and perform consistently in both test sets; meanwhile, the reconstruction results are better on different individuals when using Individual 2 as the training set.

However, in the model without using alignment and visualization, when using Individual 1 as the training set, it only performs well in its own test set, while generating completely inconsistent faces, genders, and expressions under Individual 2's test set; at the same time, the result is fainter when using Individual 2 as the training set, but it also reveals the features of the original face only under its own test set, and it works poorly under Individual 1's test set. This shows the importance of alignment and pre-training in the generalizability ability of the model.

## 6 Conclusion and Discussion

In this paper, we presented a training pipeline for cross-modal generation like reconstruction images from fMRI where image-neural data pairs are lacking but data from image domain is sufficient. Also, we proposed a novel reconstruction framework called DFGAN. The proposed DFGAN mitigates the imbalance of generating only one category of images that is too easy for the generator and too difficult for the discriminator. We also designed a pre-training process using features extracted from images as conditions that can be used in domains where neural data is lacking. Our method achieves improvements on performance and robust in the current dataset, and shows great generalization capability.